\documentclass[10pt,twocolumn,letterpaper]{article}

\usepackage{wacv}
\usepackage{times}
\usepackage{epsfig}
\usepackage{graphicx}
\usepackage{amsmath}
\usepackage{amssymb}

\usepackage[pagebackref=true,breaklinks=true,letterpaper=true,colorlinks,bookmarks=false]{hyperref}

\wacvfinalcopy 

\def\wacvPaperID{****} 

\ifwacvfinal\fi
\begin{document}

\title{Considering Race a Problem of Transfer Learning}

\author{Akbir Khan  \hspace{2cm} Marwa Mahmoud\\
The Computer Laboratory \\
University of Cambridge \\
{\tt\small ak2062@alumni.cam.ac.uk}
}
\maketitle
\ifwacvfinal\thispagestyle{empty}\fi

\begin{abstract}
As biometric applications are fielded to serve large population groups, issues of performance differences between individual sub-groups are becoming increasingly important. In this paper we examine cases where we believe race is one such factor. We look in particular at two forms of problem; facial classification and image synthesis. We take the novel approach of considering race as a boundary for transfer learning in both the task (facial classification) and the domain (synthesis over distinct datasets). We demonstrate a series of techniques to improve transfer learning of facial classification; outperforming similar models trained in the target's own domain. We conduct a study to evaluate the performance drop of Generative Adversarial Networks trained to conduct image synthesis, in this process, we produce a new annotation for the Celeb-A dataset by race. These networks are trained solely on one race and tested on another - demonstrating the subsets of the CelebA to be distinct domains for this task. 
\end{abstract}
\section{Introduction}
The adoption of machine learning technologies for commercial applications has occurred at a breathtaking rate. Nowhere has this been more clear than within the 
computer vision community, where Convolutional Neural Networks (CNNs) have revolutionised image classification \cite{rosten2006machine, Redmon:2015aa} and Generative Adverserial Networks (GANs) have pushed image synthesis to photorealistic levels \cite{DBLP:journals/corr/Wu0ZH17,DBLP:journals/corr/LedigTHCATTWS16}. Such models require an enormous amount of data to be trained robustly and are employed within services for large global user bases.

However, with a reliance on diverse data, models are unable to produce distributions which encompass all user cases - effectively restricting the audiences for these technologies. Recent attention has been brought to the inability of technologies to work well on people of certain races, for example; Asian faces are consistently classified as blinking \cite{blinking}, people of colour are whitewashed by filters \cite{white-washed} and Black people are mislabelled as gorillas \cite{buolamwini2018gender}. Whilst these are reported, we find limited studies by the research community to document such phenomena.

A common response to such reports, is to suggest models should be trained on more representative data, however such data may not exist. Even as datasets continue to grow in size, this only acts to exaggerate existing bias' within data. Furthermore, whilst selective sub-sampling of large datasets could solve this problem, we demonstrate the difficulties in this paper.

To that end, we reconsider these phenomena as failings of \textit{transfer learning}: that models are unable to generalise well beyond race in biometric tasks. 

Transfer learning considers the application of models which form a distribution of a domain/task and then are applied to another distinct domain/task. In the case of the failing technologies above, the transfer domain boundary is race, in which our models are trained on a \textit{source domain} $D_{S}$, predominately of race $X$, and then applied to a \textit{target domain}, $D_{T}$ of race $Y$.

In this paper we explore if race is a distinct transfer boundary, that is to say, a distinct characteristic of a dataset over which it is difficult to generalise. We quantify this claim through two tasks, facial classification and synthesis. For classification, we consider a series of datasets annotated by race and attempt to predict race across domains. From this task, we demonstrate the difficulty in classifying by race and, by proxy, the difficulty in selectively sub-sampling a racially diverse subset from larger datasets.

We also consider facial synthesis, the task of combining and augmenting existing facial images and artistic styles through some semantic domain. For example, a model can receive an image of a red haired women and generate a photo-realistic image of the same individual with blue hair - such tasks require a strong understanding of the identity and the features of the individual. From this task, we substantiate the problems with this specific architecture - adding weight to the claims that this technology is susceptible to bias against minority groups. 

The remainder of this paper is organised as follows: In Section 2, we attempt to apply racial classification over datasets and incorporate transfer learning techniques to improve accuracy. In Section 3, we examine the effectiveness of GANs to produce photo-realistic images over racially distinct domains. We discuss our findings in Section 4 and suggest further work in Section 5.

\section{Racial Classification}

Racial classification is a diverse topic, with approaches from classical methods to machine learning \cite{jain201650}. A large body of research exists segmenting facial images by Asian vs Non-Asian labels \cite{lyle2010soft,zarei2012artificial,qiu2006global} and finer Asian distinctions \cite{Wang:2016aa}. Fue et al. provide a full survey of techniques \cite{fu2014learning}. 

These examples choose train and test sets from the same domain, where subjects are taken in similar poses, lightening and occlusion. We explore how such a classifier performs when applied cross domain (e.g. to a different image set), and ask \textit{'Do models learn a generalised concept of race?'}. If such a classifier does exist, then racial subsampling of unseen datasets could be possible. Referencing the commercial problems discussed above, we explore the problem of Caucasian vs Non-Caucasian boundary.

Ahmed et. al. combine transfer learning with racial classification, by introducing new regularisation term based on the output of pseudo-tasks \cite{ahmed2008training}. However, these psuedo-tasks must be tailored to the datasets in use. Whilst not directly comparable, our techniques are less involved and obtain a greater accuracy boost than the comparative 0.2\% achieved here. 

\subsection{Experimental Set Up}

We trained a convolutional neural network (CNN) to classify race over a dataset of facial images. This model is then applied to an unseen domain/dataset (i.e. different facial datasets), thus testing the models ability to learn a generalised concept of race. We consider two facial datasets, the Celeb Faces Attributes Dataset (CelebA) \cite{liu2015faceattributes} and the University of Tennessee, Knoxville Face Dataset (UTKFace) \cite{zhifei2017cvpr}. These provide a controlled set of images which are racially diverse (in the case of the UKTFace, the dataset is annotated by race) but ensure individuals and their features are clearly visible. We choose the UKTFace dataset as the source domain and the CelebA dataset as the target domain.

The UKTFace dataset contains over 20,000 face images with annotations of age, gender, and race. Race is categorised into five categories: White, Black, Asian, Indian and Others. For our classification tasks, this is reduced to two categories (White, Other) with a data split of (8373, 11386) respectively. 

The CelebA dataset consists of over 200,000 face images of predominately Western celebrities. Each image comes with over 40 attributions (Age, Facial Hair, Hair Colour, Hair Style). Similar to UKTFace, images cover a large variation in pose, facial expression, illumination and occlusion. However images are not annotated by race. We note the CelebA dataset consists of individuals of different race, yet similar skin-tone (potentially due to industry beauty standards). To establish a ground truth, a set of 2500 images were annotated by two researchers, this set was further split into a training set (2000 images) and a validation set (500 images). 

\begin{figure}[h!]
\label{fig:long}
\begin{center}
   \includegraphics[width=0.8\linewidth]{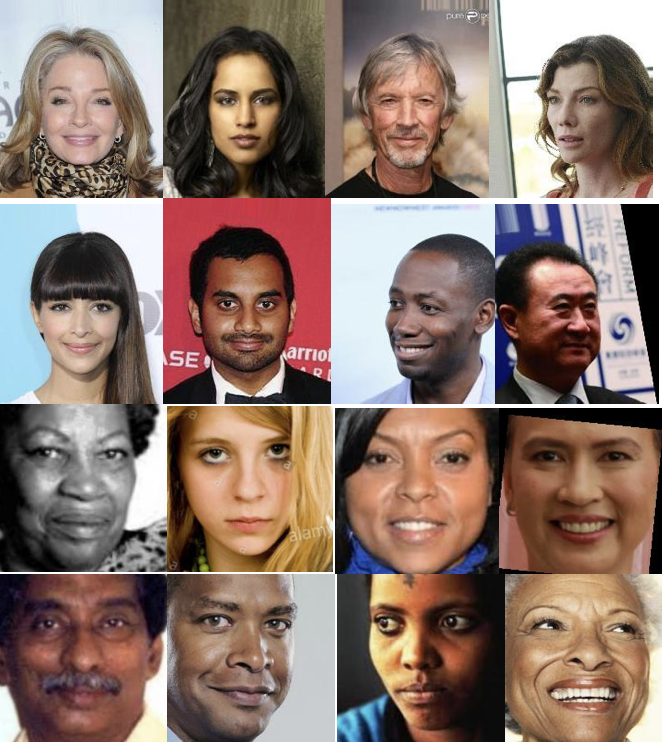}
\end{center}
   \caption{\label{Fig: Datasets}Samples of the Celeb A dataset (1st and 2nd) and UTKFace dataset (3rd and 4th Row)}

\end{figure}

Given the relatively small dataset, pre-trained neural networks are used for feature extraction. The extracted features are then fed into a Multi-layer Perceptron (MLP) consisting of 3 dense layers of 64 neurons, followed by a single output neuron. A ReLU activation function is used within the first layer and a Sigmoid function is used for the latter. We implemented a dropout = 0.5, a binary cross entropy loss function and use the Adam optimiser with learning rate = 0.001. We tested on 4 major architectures: ResNet50 \cite{He2015}, VGGFace \cite{Parkhi15}, VGG16 and VGG19 \cite{Simonyan14c}. The MLP is trained for 30 epochs with batch size 128. 


For a benchmark comparison, we evaluate a model trained within the target domain - that is to say, a model which is trained and tested purely on the annotated CelebA dataset (Train:2000, Valid:500).  

\subsection{Transfer Learning Techniques}
To improve our score we employed a series of Transfer Learning techniques found within multiple papers \cite{NIPS2012_4824,Razavian_2014_CVPR_Workshops}. 

\textbf{Fine Tuning Networks} - In order to improve feature extraction, the last 5 layers of the model were unfrozen (approximately 30\% of the model). Compared to initial training, a lower learning rate is chosen during this training. Weights were also initialised from the values obtained from initial training. 

\begin{figure}[ht]
\begin{center}
   \includegraphics[width=0.8\linewidth]{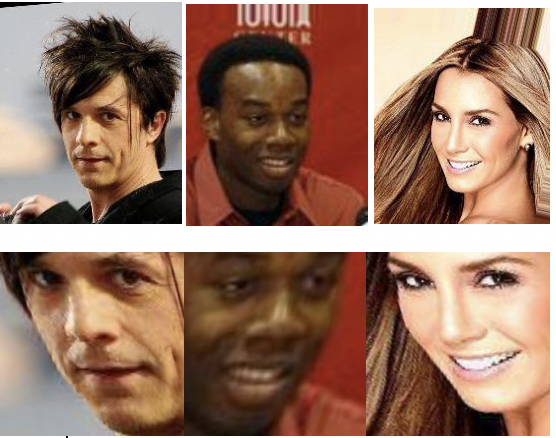}
\end{center}
   \caption{ \label{fig:agumented} Samples of images augmented by our techniques.}
\label{fig:augmented}
\end{figure}

\textbf{Data Augmentation} - To bring the domains closer together, a number of transformations are applied. Using the OpenFace library, we cropped all images to only show the face, aligning the crop with the ears and chin of an image. Images were also scaled to be the same size as the UTKFace counterparts (200 x 200). Example of transforms are presented by Figure \ref{fig:agumented}.

Further transforms included zoom, sheer and randomised shuffling. This was motivated to introduce variety to data, and to mimic the distortions within the CelebA dataset. We note that whilst augmentations can reduce the photorealism of images, the race of an individual was not obfuscated.

\textbf{Optimiser Choice} - The optimiser chosen for a model contributes largely to when a model converges and in which local minimum (of the loss function) this occurs. Models which perform poorly fall into minima which fail to encapsulate a generalisable understanding of race. We explored different optimisers, for tuning and fine-tuning. We consider three optimisers: \textit{Adam} \cite{DBLP:journals/corr/KingmaB14}, \textit{Stochastic Gradient Descent}\cite{robbins1951} and \textit{RMSprop} \cite{hinton2012neural}. 


Finally, for an accurate benchmark, we compare our augmented model with the corresponding model trained purely on the target domain, the CelebA dataset (Train: 2,000, Valid: 500).

\subsection{Results}
We report accuracy for our models on both the source and target domain. Motivated by the number of techniques to implement, we choose to optimise over individual parameters one at a time. This is similar to \textit{co-ordinate descent}, where all other variables are frozen, whilst a single variable is optimsied. Multiple repeats ($n=5$) were conducted and the average scores reported.
\begin{table}[h!]
\begin{center}
\begin{tabular}{|l|c|c|}
\hline
Model & Source & Target \\\hline\hline
ResNet 50 (Tuned)&  0.751& 0.362\\
ResNet 50 (Fine Tuned) & 1.00 & 0.386\\ \hline
VGG16 (Tuned) & 0.781& 0.421\\
VGG 16 (Fine Tuned) & 1.00 & 0.453\\ \hline
VGG19 (Tuned) &0.823& 0.383\\
VGG19 (Fine Tuned) &1.00 & 0.402\\ \hline
VGGFace (Tuned)&0.840 & 0.452\\
\textbf{VGG Face (Fine Tuned)}& \textbf{1.00} & \textbf{0.482}\\ 
\hline
\end{tabular}
\end{center}
\caption{Reported accuracy for off-the-shelf and fine-tuned models. \label{finetuned}}
\end{table}

We report untuned and fine-tuned architecture's performance upon test sets in source (UKTFace) and target (CelebA) domains (see Table \ref{finetuned}). As expected, the models perform badly, not only in the target domain but also the source domain. We note that even when models are fine-tuned, performance only improved in the source domain whilst performance dropped on the target domain. This suggests that fine-tuning models results in over-fitting.

\begin{table}[h!]
\begin{center}
\begin{tabular}{|l|c|c|c|}
\hline
Model & Unaugmented & Agumented  \\
\hline\hline
ResNet 50 &1.00/0.386  & 1.00/0.582\\
VGG16 & 1.00/0.453 & 1.000.678\\
VGG19  &1.00/0.402 & 1.00/0.621\\
VGGFace & 1.00/0.482 & 1.00/0.681\\
\hline
\end{tabular}
\end{center}
\caption{The effects of data-augmentation over a model. We report the Source/Target Domain accuracy. \label{data agument}}
\end{table}

We next applied data augmentation, presenting our results in Table \ref{data agument}. This technique increases the models accuracy on the CelebA dataset and in particular the VGG Face makes the largest improvements. From these initial results, we chose to use the VGGFace architecture for the rest of this paper.

\begin{table}[h!]
\begin{center}
\begin{tabular}{|l|c|c|c|}
\hline
Optimiser&Loss &Accuracy &Time\\
\hline \hline
Adam/Adam& 0.453 & 0.681 &24\\
Adam/SGD &0.446 & 0.721&26\\
RMSprop/SGD&0.291 & 0.910&45\\
SGD/RMSprop&0.311 & 0.871&44\\
\hline
\end{tabular}
\end{center}
\caption{Results for different optimisers. Time is measured in Epochs till Convergence. \label{Table: Optimiser}}
\end{table}

We report differing performance based on the optimiser used. For all prior experiments, the Adam optimiser was utilised. Furthermore, we also report the number of epochs till these models converge. We present out findings in Table \ref{Table: Optimiser}.

We finally report the performance of in-domain trained models (e.g. both source and target domain are the same) against our out-of-domain trained model. Our results are presented in Table \ref{final}. The target domain model scored $0.83$ on the validation set, whilst our transfer learning model achieves $0.91$. This crucially demonstrates that whilst from initial findings, racial classifiers fail to obtain generaliable concepts of race - employing transfer learning methods results in models with better performance scores and thus a more generalised concept of race. 

\begin{table}[ht]
\begin{center}
\begin{tabular}{|l|c|c|}
\hline 
Model& Source & Target\\ \hline
Benchmark &1.00 &0.83  \\ \hline
Baseline &{0.840}&{0.452} \\ \hline
Tuned &{1.00}&{0.482} \\ \hline 
Tuned + DataAugmentation&1.00&0.621 \\ \hline
\textbf{Tuned + PreProcessing + Optimisers}& \textbf{1.00} & \textbf{0.91} \\ \hline
\end{tabular}
\end{center}
\caption{Models trained in different domains outperforms those trained in the target domain. Benchmark refers to a model trained in the target domain whilst Baseline is the vanilla VGGFace network. \label{final}}
\end{table}

\section{Image Synthesis}
Our second task focused upon image synthesis from Generative Adversarial Networks (GANs). We test the current state of the art solution for photorealistic image generation, the StarGAN network, and its ability to generate synthesised images across domain \cite{Choi:2017aa}. 

We explore the importance of a racially balanced dataset for such a task, or more specifically \textit{'Is race a characteristic to distinguish transfer learning domains?'}. We evaluate how a model, which is trained to conduct image synthesis upon a dataset of a specific race (source domain), performs when applying the same synthesis on a dataset of another race (target domain). Performance is evaluated by the photo-realism of generated images. 

Note that both domains may be within the same dataset (e.g. the CelebA dataset). If race were a trivial distinction, performance would be unaffected by which set we use to train on; our goal is to demonstrate this is not the case. To our knowledge there is no literature on cross-domain application of GANs. Although GANs have been used to help train cross-domain classifiers \cite{Yoon:2017aa}, we believe this is a novel study.

\subsection{Dataset}
We chose to utilise the CelebA dataset for its wide range of labels, which allow for multiple permutations of images by facial feature synthesis. In addition, the performance of the StarGAN architecture is well reported on the CelebA dataset. 

From work demonstrated in Section 2 and further human annotation, we segmented the CelebA dataset into two distinct subsets: CelabC - a strictly caucasian dataset and CelebO - containing all other races (140638, 56375 entries respectively). Given the prominence of individuals within the CelebA dataset, annotations were easily verified (by image search) and any contentious cases were placed in CelebO. Example of contentious images are provided in Figure \ref{fig:contentious}.

\begin{figure}[ht]
\begin{center}
   \includegraphics[width=0.8\linewidth]{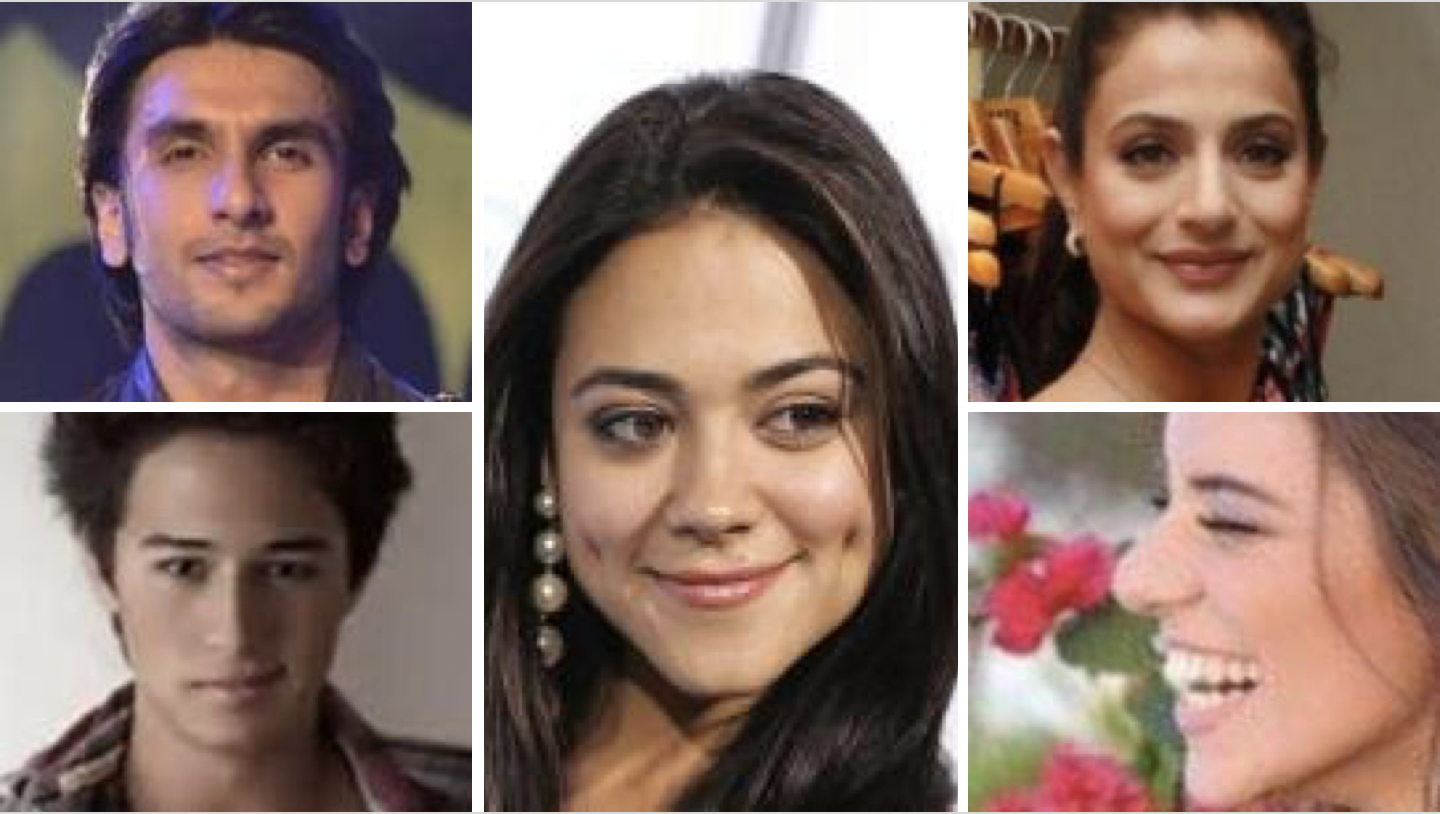}
\end{center}
   \caption{ \label{fig:contentious} Samples of contentious images, placed in Celeb-O}
\label{fig:contentious}
\end{figure}

\subsection{Image Generation}
GANs are composed of two competing neural networks: a \textit{Generator} $G$, which creates images and a \textit{Discriminator} $D$, a network which classifies if the image is real or fake. Both components are trained together, optimising to outperform each other and thus producing a photorealistic generator.

In the StarGAN architecture, the \textit{Generator} is a function of an initial image and an \textit{attribute vector}, with the goal to synthesise the image with a desired facial feature (or attribute). This image is then passed to a \textit{Discriminator} which learns to distinguish real and fake images and predict the augmented feature.

The architecture has the ability to have multiple attributes/labels. We choose to apply synthesis over single attributes, synthesing new images with changed age, facial hair, hair colour or hair style. The architecture currently stands as state of the art for the photorealism it achieves. This is in particular due to the introduction of a Reconstruction Loss, in addition to the vanilla domain classification loss. For the purpose of our experiments, we wish to leverage the photorealistic capabilities of the StarGAN to highlight performance drops on facial synthesis technologies on minority groups.

\begin{figure*}[h!]
\begin{center}
\includegraphics[width=0.75\textwidth]{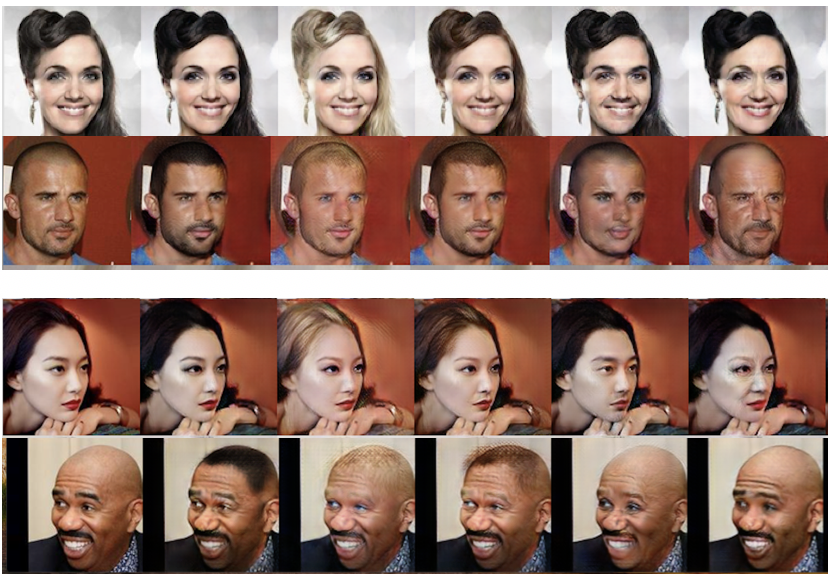}
\end{center}
  \caption{Images synthesis from the CelebC and CelebO datasets (Top/Bottom resp.) \label{generated}}

\end{figure*}

\subsection{Training}

Models were trained for over 20 epochs with a batch size 16 and 200,000 iterations per image. Models were trained with the Adam optimiser with $\beta_{1}=0.5$ and $\beta_{2}=0.999$. All images were cropped to a size of 178 x 178 and a learning rate of 0.0001 was used for both generator and discriminator. We performed one generator update for every five discriminator updates. A model's training phase was completed within 26 hours. This process was repeated with 5 different random initialisations and all generated images were collected. 

\subsection{Evaluation}

For testing, two output sets were produced. These sets contain images which are generated with a set of features (e.g. Hair Colour and Age) changed from an original image. The first set of outputs is generated from an unseen portion of the CelebC dataset (in-domain), whilst the second is generated from the CelebO dataset (out-domain). All our experiments were conducted using the model output from images unseen during the training phase. To evaluate the performance we explore both qualitative and quantitative evaluation.

\textbf{Qualitative} - Figure \ref{generated} shows a sample of the facial attribute transfer on both CelebC and CelebO. Transforms were least realistic for males from the CelebO dataset. We postulate that, by examining the two datasets, these individuals have the least physical resemblance to members of the CelebC dataset. Based on this, it appears comparison women fare significantly better.

\textbf{Quantitative} - A series of double blind test was applied with a group of participants \footnote{This experiment was conducted with the approval of Cambridge Computer Science Ethics Committee}. 
Given a pair of images (one from each output set), participants were instructed to choose the best generated
image based on perceptual realism, quality of transfer in attribute(s), and preservation of the identity of the original figure. Our study was taken by 3 participants viewing a total of 300 pairs each. Given our qualitative evaluation we also controlled for gender within each image.
\begin{table}[h!]
\begin{center}
\begin{tabular}{|c|c|c|}
\hline
Gender& CelebC & CelebO\\
\hline \hline
Male& 60\%& 40\% \\
Female &53\% & 47\%\\
\hline
\end{tabular}
\end{center}
\caption{Percentage of participants which found an image from a specific dataset photo-realistic.}
\end{table}

From this, we found the model performed worse on Celeb-O dataset, and thus performed worse on non-caucasian faces, agreeing with the qualitative observations. We believe it is clear that the photo-realism of the fake images is worsened when applied to images from a different race.

\section{Discussion}

The results presented in this paper contribute to understanding if race can be considered a problem of transfer learning. In the case where this has been proven true, the effect of an unbalanced dataset become cruical. Such a fact could be considered \textit{a priori} true given the large data dependency of machine learning. However, our study highlights a tangible evaluation of how poorly these models will perform in said scenario, for those specific underrepresented subgroups. 

For Section 3, we acknowledge that the results of the study will adopt the biases of the participants involved. As shown by Johnston et al. \cite{johnston2009familiar}, humans process familiar and unfamiliar faces differently. Thus, we postulate that the synthesised image of a familiar face vs unfamiliar is also processed differently and could alter the photorealistism percieved by candidates. 

This bias becomes more apparent when, through an additional questionnaire, participants reported they were aware of 85\% of all celebrities listed in the CelebC but only 65\% of all those listed in the CelebO dataset. This suggests that the reported discrepancy in photorealism would be larger in a commercial setting (with less famous users). Thus, we consider our results to be a lower-bound on the performance difference between image synthesis over different races.

With the problem well defined, we now consider potential solutions. As aforementioned, a common suggestion is selectively subsampling larger datasets to create a balanced training set. Whilst we report strong results on such classifiers in Section 2, the level of hyperparameter optimisation between differing datasets is intensive, bringing the commercial viability into question. We posit that further work is  required before for such a class of solution is viable.

\section{Conclusion}
In conclusion, we provide three main contributions within this paper. We firstly highlight effective techniques to improve racial classifiers of facial images. These techniques produce a model which, whilst trained in a separate domain, scores higher than benchmarks trained in the same domain. This demonstrates the effectiveness of transfer learning in problems of race classification. It also highlights the difficulty within subsampling racially diverse datasets from large unbalanced datasets.

We secondly produce new annotations for the CelebA dataset, a set of 200,000 images with with an additional label (Race). Thirdly, we demonstrate that race plays a significant factor in a generative models' ability to produce photo-realistic images.

We note that race is a difficult characteristic to identify in a cross-domain learning environment. Although the results from our first experiment are significant improvements, model performance scores are still poor. Similarly, we believe that our second experiment further highlights the real distinction race can play in image generation. We hope that this study can further highlight the importance of racially diverse data to those utilising biometric technologies

{\small
\bibliographystyle{ieee}
\bibliography{egpaper_final}
}

\end{document}